\documentclass[journal,9pt]{IEEEtran}
\usepackage[utf8]{inputenc}
\ifCLASSINFOpdf
\else
\usepackage[dvips]{graphicx}
\fi
\usepackage{url}
\usepackage{multirow}
\usepackage{bm}
\usepackage{graphicx}
\usepackage{amsfonts}
\usepackage{amsmath}
\usepackage{booktabs}
\usepackage{tabularx}
\usepackage{amssymb}
\usepackage{caption}
\usepackage{float}
\usepackage{xcolor}
\usepackage{cite}
\usepackage[a4paper, left=1.0cm, right=1.0cm, top=1.0cm, bottom=0.8cm]{geometry}
\usepackage{cite}
\usepackage{balance}
\usepackage{hyperref}
\begin{document}
	
	\title{DWAFM: Dynamic Weighted Graph Structure Embedding Integrated with Attention and Frequency-Domain MLPs for Traffic Forecasting}
	\author{Sen~Shi, Zhichao~Zhang,~\IEEEmembership{Member,~IEEE}, and Yangfan~He
		\thanks{This work was supported in part by the Open Foundation of Hubei Key Laboratory of Applied Mathematics (Hubei University) under Grant HBAM202404; in part by the Foundation of Key Laboratory of System Control and Information Processing, Ministry of Education under Grant Scip20240121; and in part by the Startup Foundation for Introducing Talent of Nanjing Institute of Technology under Grant YKJ202214. \emph{(Corresponding author: Zhichao~Zhang.)}}
		\thanks{Sen~Shi is with the School of Mathematics and Statistics, Nanjing University of Information Science and Technology, Nanjing 210044, China (e-mail: 202312380032@nuist.edu.cn).}
		\thanks{Zhichao~Zhang is with the School of Mathematics and Statistics, Nanjing University of Information Science and Technology, Nanjing 210044, China, with the Hubei Key Laboratory of Applied Mathematics, Hubei University, Wuhan 430062, China, and also with the Key Laboratory of System Control and Information Processing, Ministry of Education, Shanghai Jiao Tong University, Shanghai 200240, China (e-mail: zzc910731@163.com).}
		\thanks{Yangfan~He is with the School of Communication and Artificial Intelligence, School of Integrated Circuits, Nanjing Institute of Technology, Nanjing 211167, China, and also with the Jiangsu Province Engineering Research Center of IntelliSense Technology and System, Nanjing 211167, China (e-mail: Yangfan.He@njit.edu.cn).}}
	\markboth{}
	{Shell \MakeLowercase{\textit{et al.}}: Bare Demo of IEEEtran.cls for IEEE Journals}
	\maketitle
	
	\begin{abstract}
		Accurate traffic prediction is a key task for intelligent transportation systems. The core difficulty lies in accurately modeling the complex spatial-temporal dependencies in traffic data. In recent years, improvements in network architecture have failed to bring significant performance enhancements, while embedding technology has shown great potential. However, existing embedding methods often ignore graph structure information or rely solely on static graph structures, making it difficult to effectively capture the dynamic associations between nodes that evolve over time. To address this issue, this letter proposes a novel dynamic weighted graph structure (DWGS) embedding method, which relies on a graph structure that can truly reflect the changes in the strength of dynamic associations between nodes over time. By first combining the DWGS embedding with the spatial-temporal adaptive embedding, as well as the temporal embedding and feature embedding, and then integrating attention and frequency-domain multi-layer perceptrons (MLPs), we design a novel traffic prediction model, termed the DWGS embedding integrated with attention and frequency-domain MLPs (DWAFM). Experiments on five real-world traffic datasets show that the DWAFM achieves better prediction performance than some state-of-the-arts.
	\end{abstract}
	
	
	\IEEEpeerreviewmaketitle
	
	\section{Introduction}
	\IEEEPARstart{T}{raffic} forecasting \cite{1} is a spatial-temporal sequence prediction task that infers future traffic states (such as flow and speed) based on historical spatial-temporal observational data. In recent years, deep learning has made significant progress in this field. Spatial-temporal graph neural networks (STGNNs) \cite{2,3,4,5,6} and models based on attention mechanisms \cite{7,8,9,10} have attracted considerable attention due to their remarkable ability to effectively capture complex spatial-temporal dependencies. In pursuit of higher prediction accuracy, research has focused on designing more complex model architectures, such as novel graph convolutions \cite{11,12,13,14,15,16} and learning graph structures \cite{17,18,19,20,21,22,23}. However, researchers have gradually observed that improvements in network architecture often fail to significantly enhance model performance. This phenomenon has prompted a shift in research from complex architectures to exploring data representation techniques.
	
	In this context, input embedding technology, as a critical component of data representation, has demonstrated tremendous potential and become a research hotspot. Specifically, Shao et al. proposed STID, which achieves excellent performance while maintaining high efficiency by introducing independent spatial-temporal identity embedding and combining it with a multi-layer perceptron (MLP) architecture \cite{24}. Liu et al. proposed STAEformer, which designs a spatial-temporal adaptive embedding component to significantly improve the representation ability of spatial-temporal information \cite{25}. Meanwhile, Wang et al. embedded predefined static graph structure information into the spatial representation in their ST-MLP research, further enhancing the performance of the model in spatial-temporal prediction tasks \cite{26}. These successful practices highlight the effectiveness of carefully designed embedding techniques in simplifying model architectures while enhancing performance. However, these methods either completely avoid using predefined graph structures, leading to insufficient spatial relationship modeling, or introduce predefined static graph structures that fail to accurately reflect dynamic changes in node association strength during scenarios like morning and evening rush hours.
	
	To overcome this critical limitation, this letter proposes an innovative dynamic weighted graph structure (DWGS) embedding method. While preserving the predefined graph topology, this method adaptively learns the time-varying weights of each edge in the graph in a spatial-temporal data-driven manner, thereby more accurately characterizing the dynamic spatial dependencies implicit in spatial-temporal datasets. Furthermore, considering the potential limitations of such a DWGS in representing complex spatial relationships between nodes, we introduce a spatial-temporal adaptive embedding to capture the underlying spatial correlation patterns in the data. Overall, by combining the DWGS embedding with the spatial-temporal adaptive embedding, we construct a unified and more expressive spatial representation, enhancing the ability of modeling complex spatial-temporal dependencies.
	
	Ultimately, this letter proposes a novel traffic prediction model called the DWGS embedding integrated with attention and frequency-domain MLPs (DWAFM), by integrating the collaborative working mechanism of attention and
	frequency-domain MLPs, combined with the temporal embedding and feature embedding. This model jointly learns complex spatial-temporal dependencies to achieve high-performance traffic prediction. The main contributions of this study are summarized as follows:
	\begin{itemize}
		\item A DWGS embedding method is proposed, where the dynamic weighted adjacency matrix is learned adaptively by data-driven way to ‌characterize the change of the correlation strength between nodes over time.
		\item A novel traffic prediction model, the DWAFM, is constructed, effectively combining the advantages of self-attention in spatial relationship modeling and frequency-domain MLPs in temporal pattern extraction, achieving efficient modeling of complex spatial-temporal dependencies.
	\end{itemize}

	\section{Preliminaries}
	
	\textit{Definition  1. Traffic Network:} The traffic network is modeled as an undirected graph $\bm{G} = (\bm{V}, \bm{E}, \mathbf{A})$ that maps the physical road infrastructure into a topological space, where the vertex set $\bm{V} = \{v_1, v_2, \dots, v_N\}$ represents $N$ discrete observation points corresponding to physical sensors (e.g., loop detectors or surveillance cameras) deployed at road segments or intersections to record continuous time-series data. The edge set $\bm{E}$ defines connectivity based on the physical road layout, where an edge $e_{ij}$ exists if and only if there is a direct navigable road segment between observation locations, while the adjacency matrix $\mathbf{A} \in \mathbb{R}^{N \times N}$ characterizes the spatial correlation within the network, with each entry $\mathbf{A}_{i,j}$ quantifying the degree of association---such as spatial proximity or historical traffic correlation---between locations if a physical connection exists, and $0$ otherwise.
	
	\textit{Definition  2. Traffic Forecasting:} The goal of traffic forecasting  is to use traffic data \(\mathbf{X} \in \mathbb{R}^{T \times N \times D}\) from the past \(T\) time periods to predict data \(\mathbf{Y} \in \mathbb{R}^{T_{f} \times N \times D}\) for the next \(T_f\) time periods. This traffic prediction task is defined as
	\[\mathbf{X} \xrightarrow{F} \mathbf{Y},\]
	where \(F\) represents a learning function that maps historical data to future data, and \(D\) is the dimension of traffic data features.
	
	\begin{figure*}[ht]
		\centerline{\includegraphics[width=\textwidth]{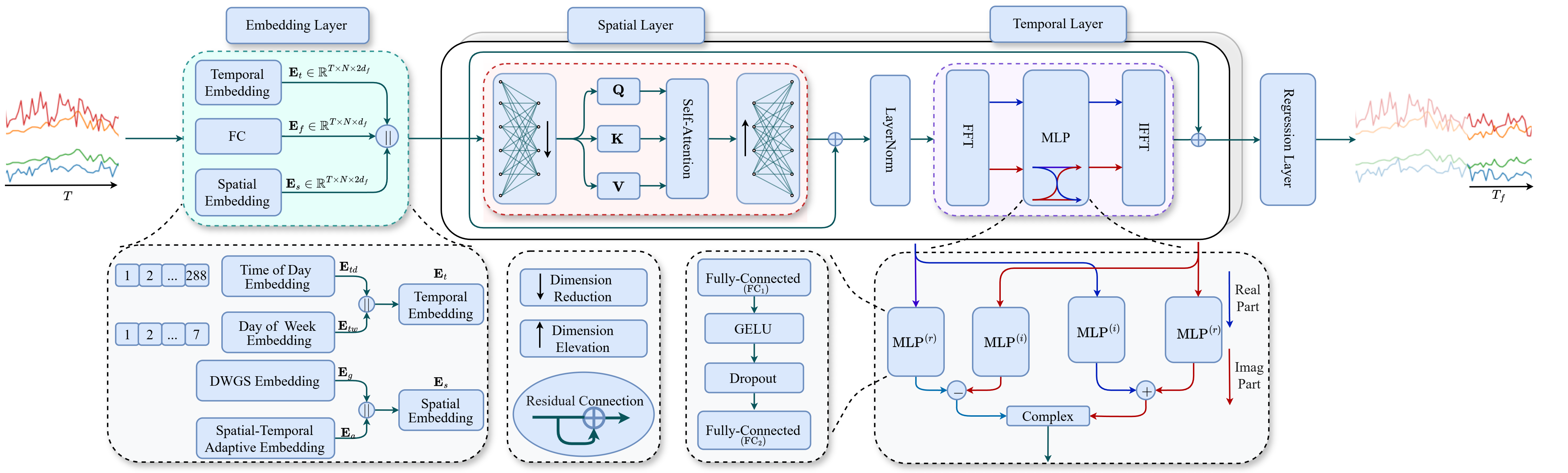}}
		\caption{The overview of the proposed DWAFM.}
		\label{fig:1}
	\end{figure*}
	\section{Methodology}
	\label{sec:guidelines}
	
	\subsection{Overview Framework}
	As shown in Fig.~\ref{fig:1}, our model consists of an embedding layer, a spatial layer, a temporal layer, and a regression layer. The embedding layer fuses feature embedding, temporal embedding, and spatial embedding to generate a hidden representation. The spatial layer and the temporal layer are used to capture spatial relationships between traffic sequences and temporal dependencies within sequences, respectively. The regression layer ultimately completes the prediction.
	
	\subsection{Embedding Layer}
	
	\begin{figure}
		\centerline{\includegraphics[width=\linewidth]{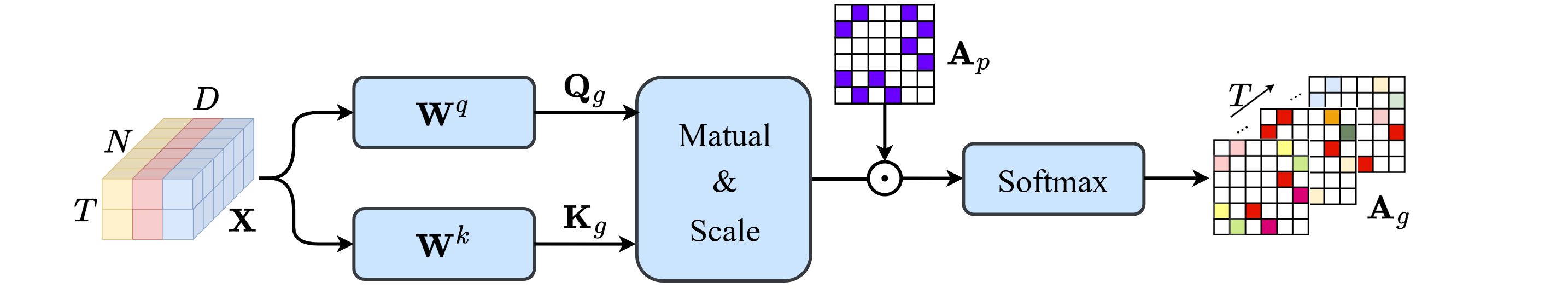}}
		\caption{Learning mechanism of dynamic weighted adjacency matrix.}
		\label{fig:2}
	\end{figure}
	
	The embedding layer primarily serves to map raw input data into high-dimensional space, enabling better capture of complex spatial-temporal dependencies. To preserve the feature information from the original data \( \mathbf{X}\), it is first translated into a feature embedding:
	\[\mathbf{E}_f = \text{FC}(\mathbf{X}) \in \mathbb{R}^{T \times N \times d_f},\]
	where \(\text{FC}(\cdot)\) denotes a fully-connected layer, and \( d_f\) represents the embedding dimension.
	
	For temporal embedding, two types of learnable representations are defined to capture periodic and trend characteristics in traffic data: \( \mathbf{W}_d \in \mathbb{R}^{N_d \times d_f}\) for encoding intraday patterns and \( \mathbf{W}_w \in \mathbb{R}^{N_w \times d_f}\) for encoding intraweek patterns, where \(N_d\) and \(N_w\) denote the number of time intervals in a day and days in a week, respectively. For each time step \(t\), the corresponding embedded vectors derive from these representations based on their specific temporal positions. Since all traffic nodes share the same temporal characteristics, the retrieved time-of-day and day-of-week embeddings are replicated and expanded along the node dimension, giving birth to \(\mathbf{E}_{td} \in \mathbb{R}^{T \times N \times d_f}\) and \(\mathbf{E}_{tw} \in \mathbb{R}^{T \times N \times d_f}\). Then, \(\mathbf{E}_{td}\) and \(\mathbf{E}_{tw}\) are concatenated to form the temporal embedding
	\[\mathbf{E}_t = \text{Concat}\left[(\mathbf{E}_{td}, \mathbf{E}_{tw}), \text{dim} = -1 \right] \in \mathbb{R}^{T \times N \times 2d_f}.\]
	
	For spatial embedding, traditional graph structure embedding methods typically rely on predefined adjacency matrices, which fail to reflect the dynamic strength of associations between nodes. Therefore, we propose a dynamic weighted adjacency matrix, built upon existing predefined adjacency matrices, to capture the time-varying strength of associations between nodes.
	As shown in Fig.~\ref{fig:2}, we employ a self-attention on the original data \( \mathbf{X}\) to learn the temporal dynamics of the connection weights. This process is defined as
	\[\mathbf{Q}_g = \mathbf{X} \mathbf{W}^q,  \mathbf{K}_g = \mathbf{X} \mathbf{W}^k,  \mathbf{A}_a = \text{Softmax}\left(\frac{\mathbf{Q}_g \mathbf{K}_g^\top}{\sqrt{d_f}}\odot \mathbf{A}_p\right),\]
	where \(\mathbf{W}^q,\mathbf{W}^k \in \mathbb{R}^{D \times d_f}\) are learnable parameters, and \(\mathbf{A}_p\) is a predefined adjacency matrix. To enforce zero attention scores at non-connected positions of \(\mathbf{A}_p\), we mask the corresponding positions in the attention logits by setting them to negative infinity prior to applying Softmax. \(\mathbf{A}_a\) represents the attention scores, which is asymmetric. To construct a valid adjacency matrix, we symmetrize \(\mathbf{A}_a\) to obtain the dynamic weighted adjacency matrix
	\(\mathbf{A}_g = (\mathbf{A}_a + \mathbf{A}_a^\top)/{2}\in \mathbb{R}^{T \times N \times N}.\) Subsequently,  a learnable matrix \(\mathbf{B} \in \mathbb{R}^{N \times d_f}\)  is multiplied by \(\mathbf{A}_g\) to obtain the DWGS embedding
	\[\mathbf{E}_g = \mathbf{A}_g \mathbf{B} \in \mathbb{R}^{T \times N \times d_f}.\]
	
	Considering that relying solely on \(\mathbf{E}_g\) may be insufficient to capture complex spatial relationships between nodes, we introduce a spatial-temporal adaptive embedding \(\mathbf{E}_a \in \mathbb{R}^{T \times N \times d_f}\) to capture the hidden spatial relationships, further enriching the spatial representation. Then, \(\mathbf{E}_g\) and \(\mathbf{E}_a\) are concatenated to form the spatial embedding
	\[\mathbf{E}_s = \text{Concat}\left[(\mathbf{E}_g, \mathbf{E}_a), \text{dim} = -1 \right] \in \mathbb{R}^{T \times N \times 2d_f}.\]
	By concatenating all the above embedding vectors, we obtain the final hidden spatial-temporal representation
	\[\mathbf{Z} = \text{Concat}\left[(\mathbf{E}_f, \mathbf{E}_s,\mathbf{E}_t), \text{dim} = -1 \right] \in \mathbb{R}^{T \times N \times d_h}.\]
	The hidden dimension \(d_h\) is set to \(5d_f\).
	\subsection{Spatial Layer}\label{formats}
	Inspired by the self-attention \cite{27}, we introduce an efficient spatial dependency modeling module based on a three-stage architecture of dimension reduction, self-attention and dimension elevation. This design effectively  captures complex spatial relationships between sequences while maintaining computational efficiency. 
	Given a hidden spatial-temporal representation \(\mathbf{Z}\), we first apply a dimension reduction step using two 1D convolutional layers (1D-CNNs) to compress \(T \times d_h\) dimensions into a node-level representation 
	\[\mathbf{Z}_r = \text{Conv1d}(\text{ReLU}(\text{Conv1d}(\mathbf{Z}))) \in \mathbb{R}^{N \times d_h}.\]
	We then perform self-attention on \(\mathbf{Z}_r\). The scaling factor is set to \(\frac{1}{\sqrt{{d_h}/{2}}}\), a configuration validated through hyperparameter analysis to achieve optimal performance. This process is as follows:
	\[\mathbf{Q} = \mathbf{W}_q \mathbf{Z}_r,  \mathbf{K} = \mathbf{W}_k \mathbf{Z}_r,  \mathbf{V} = \mathbf{W}_v \mathbf{Z}_r,\] 
	\[\mathbf{Z}_s = \text{Softmax}\left(\frac{\mathbf{Q} \mathbf{K}^\top}{\sqrt{{d_h}/{2}}}\right) \mathbf{V},\]
	where \(\mathbf{W}_q,\mathbf{W}_k,\mathbf{W}_v \in \mathbb{R}^{d_h \times d_h}\) are learnable parameters. The resulting attention output \(\mathbf{Z}_s \in \mathbb{R}^{N \times d_h}\) is then upsampled back to the original spatial-temporal dimensions via two additional 1D-CNNs, giving birth to
	\[\mathbf{Z}_e = \text{Conv1d}(\text{ReLU}(\text{Conv1d}(\mathbf{Z}_s))) \in \mathbb{R}^{T \times N \times d_h}.\]
	Finally, we incorporate a residual connection followed by layer normalization to produce the input representation \(\mathbf{Z}_t\) for the temporal layer. That is
	\[\mathbf{Z}_t = \text{LayerNorm}(\mathbf{Z}_e+\mathbf{Z}) \in \mathbb{R}^{T \times N \times d_h}.\]
	This structured approach enables effective spatial feature extraction with reduced computational overhead.

	\subsection{Temporal Layer and Regression Layer}
	For traffic forecasting, capturing temporal dependencies is typically accomplished using recurrent neural networks \cite{2}, \cite{28}, temporal convolutional networks \cite{3,4}, or attention mechanisms \cite{7}, \cite{29}. Recent research \cite{30} has shown that frequency-domain MLPs offer significant advantages in modeling temporal relationships. Inspired by this idea, this letter improves the frequency-domain MLPs architecture by using independent MLP modules to cross-process the real and imaginary parts of the complex spectrum, effectively capturing frequency-domain features. Specifically, we apply fast Fourier transform (FFT, denoted by \(\mathcal{F}\)) along the temporal dimension of \(\mathbf{Z}_t\) to obtain the complex representation \(\mathbf{Z}_f=\mathcal{F}(\mathbf{Z}_t)\). The real and imaginary components, \(\mathbf{Z}_f^r\) and \(\mathbf{Z}_f^i\), are then processed via parameter-independent MLPs in a cross-computation manner, yielding
	\[\hat{\mathbf{Z}}_f = \text{MLP}^{(r)}(\mathbf{Z}_f^r) - \text{MLP}^{(i)}(\mathbf{Z}_f^i) 
	+ j \Big[ \text{MLP}^{(i)}(\mathbf{Z}_f^r) + \text{MLP}^{(r)}(\mathbf{Z}_f^i) \Big],\]
	which better adapts the complex structure of frequency-domain components and effectively captures key temporal patterns with a global perspective. The structure of each MLP is based on the MLP-Mixer \cite{31} and is defined as
	\[\text{MLP}(\mathbf{Z}^{r(i)}_f) = \text{FC}_2\big(\text{Dropout}(\text{GELU}(\text{FC}_1(\mathbf{Z}^{r(i)}_f)))\big).\]
	Subsequently, the inverse FFT (IFFT, denoted by \(\mathcal{F}^{-1}\)) is applied to \(\hat{\mathbf{Z}}_f\) to obtain the result \(\hat{\mathbf{Z}}_t = \mathcal{F}^{-1}(\hat{\mathbf{Z}}_f) \in \mathbb{R}^{T \times N \times d_h}\) of the temporal layer. 
	
	Finally, the regression layer produces the prediction \(\ \hat{\mathbf{Y}} \in \mathbb{R}^{T_f \times N}\) from the final output \( {\mathbf{Z}}_{st} \in \mathbb{R}^{T \times N \times d_h}\), which is obtained through the spatial and temporal layers. The regression layer is defined as
	\[ \hat{\mathbf{Y}}= \text{FC}({\mathbf{Z}}_{st}).\]
	Here, the fully-connected layer \(\text{FC}(\cdot)\) reduces the dimension \(T \times d_h\) in \({\mathbf{Z}}_{st}\) to the dimension \(T_f \times 1\) in \(\hat{\mathbf{Y}}.\) 
	
	\section{Experiment Results and Analysis}
	\label{sec:guidelines}
	
	\subsection{Experimental Settings}
	\textit{Datasets:} We conduct experiments on five real-world traffic datasets, including three highway traffic flow datasets (PEMS03, PEMS04, and PEMS08) and two traffic speed datasets (PEMSD7(L) and PEMSD7(M)). Details of these datasets are provided in \textit{Appendix A of the Supplementary Material}.

	\textit{Baselines:} This study uses a variety of publicly available benchmark methods for comparison, including traditional methods (HI \cite{32}), typical deep learning methods (STGCN \cite{4}, GWNet \cite{3}, StemGNN \cite{17}, STNorm \cite{33}, STID \cite{24}), and some of the latest methods proposed in the past two years (DGCRN \cite{21}, MegaCRN \cite{34}, STWave \cite{35}, STAEformer \cite{25}, DFDGCN \cite{36}). Here, all of the methods are implemented on the BasicTS platform \cite{37} to ensure fair and consistent evaluations.
	
	\textit{Metrics:} To evaluate the performance of the model, several commonly used regression evaluation metrics are considered, including mean absolute error (MAE), root mean square error (RMSE), and mean absolute percentage error (MAPE(\%)).
	
	\textit{Model Settings:} Experiments are conducted under a computer environment with one Intel(R) Core(TM) i5-12400F CPU and one NVIDIA RTX 4060Ti GPU card. The model is trained using the Adam optimizer, with an initial learning rate of 0.001 and a batch size of 64. The training epoch is 80. The five datasets are divided into training, validation, and test sets in the ratio of 6:2:2. For prediction, we set the length of both input \(T\) and output \(T_f\) to 12 steps. The detailed model parameters and code are publicly available at: \href{https://github.com/ssnuist/DWAFM}{https://github.com/ssnuist/DWAFM}.
	
	\subsection{Forecasting Results}
	The model performance comparison results are shown in Table~\ref{tab:1}, where the average performance of each method across 12 consecutive time slots is listed. The best results are highlighted in bold, and the second-best results are marked with underlines. DWAFM achieves superior performance across most metrics in five datasets. Particularly on PEMS08 and PEMSD7(M), it outperforms all state-of-the-art STGNNs and transformer-based methods across all evaluation metrics. Note that while DWAFM occasionally yields slightly higher MAPE on certain datasets, this stems from the inherent mathematical sensitivity of MAPE to extremely small ground-truth values (e.g., late-night low traffic). Since our MAE-based training objective prioritizes accurately fitting peak-hour traffic trends, minor absolute deviations during low-flow periods are disproportionately amplified by the MAPE calculation. Furthermore, some baseline methods (e.g., STAEformer) also exhibit competitive performance.
	
	\begin{table*}[h]
		\centering
		\caption{Performance Comparison of Baselines and Proposed DWAFM on Five Popular used Real-world Traffic Datasets.}
		\renewcommand{\arraystretch}{0.9}
		\label{tab:1}
		\tiny
		\resizebox{\textwidth}{!}{
			\begin{tabular}{ccccccccccccccc}
				\hline Datasets & Metric & HI & STGCN & GWNet & StemGNN & STNorm & STID & DGCRN & MegaCRN & STWave & STAEformer & DFDGCN & \textbf{DWAFM}\\ \hline
				
				\multirow{3}{*}{PEMS08} 
				& MAE  & $34.66$ & $16.86$ & $14.42$ & $16.08$ & $15.47$ & $14.21$ & $14.90$ & $14.87$ & $13.78$ & $\underline{13.69}$ & $14.21$ & $\mathbf{13.57}$\\ 
				& RMSE & $50.54$ & $26.10$ & $23.38$ & $25.49$ & $25.13$ & $23.36$ & $23.76$ & $23.87$ & $23.68$ & $\underline{23.06}$ & $23.45$ & $\mathbf{22.94}$\\
				& MAPE & $21.63$ & $11.62$ & $9.33$ & $10.19$ & $9.93$ & $9.38$ & $10.17$ & $9.58$ & $9.36$ & $\underline{9.03}$ & $9.37$ & $\mathbf{8.92}$\\
				\hline
				\multirow{3}{*}{PEMS04} 
				& MAE & $42.35$ & $19.89$  & $18.76$ & $21.01$ & $18.90$ & $18.46$ & $18.96$ & $18.83$ & $18.28$ & $\underline{18.21}$ & $18.62$ & $\mathbf{18.14}$\\ 
				& RMSE & $61.66$ & $31.59$ & $30.19$ & $32.96$ & $30.50$ & $\underline{29.98}$ & $30.77$ & $30.54$ & $\underline{29.98}$ & $30.57$ & $30.01$ & $\mathbf{29.72}$\\
				& MAPE & $29.22$ & $13.61$ & $13.06$ & $14.51$ & $13.00$ & $12.55$ & $12.87$ & $12.83$ & $\underline{12.25}$ & $\mathbf{12.21}$ & $12.94$ & $12.37$ \\
				\hline
				\multirow{3}{*}{PEMS03} 
				& MAE & $32.62$ & $15.99$ & $14.63$ & $15.62$ & $15.20$ & $15.47$ & $\mathbf{14.53}$ & $14.68$ & $14.93$ & $15.31$ & $\underline{14.57}$ & $14.81$\\ 
				& RMSE & $49.89$ & $27.78$ & $25.88$ & $26.36$ & $25.94$ & $26.77$ & $25.75$ & $26.24$ & $26.04$ & $26.65$ & $\underline{24.91}$ & $\mathbf{23.96}$\\ 
				& MAPE & $30.60$ & $15.36$ & $\underline{14.59}$ & $15.03$ & $\mathbf{14.55}$ & $16.37$ & $14.97$ & $14.91$ & $15.40$ & $15.45$ & $14.89$ & $15.66$ \\ 
				\hline
				\multirow{3}{*}{PEMSD7(M)} 
				& MAE  & $5.02$ & $2.73$ & $2.67$ & $2.91$ & $2.61$ & $2.61$ & $2.70$ & $2.61$ & $2.61$ & $\underline{2.60}$ & $2.77$ & $\mathbf{2.59}$ \\
				& RMSE  & $9.58$ & $5.47$ & $5.39$ & $5.71$ & $5.34$ & $5.35$ & $5.53$ & $5.42$ & $5.33$ & $\underline{5.31}$ & $5.71$ & $\mathbf{5.24}$ \\
				& MAPE  & $12.31$ & $6.97$ & $6.79$ & $7.35$ & $6.59$ & $6.63$ & $6.87$ & $\underline{6.47}$ & $6.57$ & $6.53$ & $7.00$ & $\mathbf{6.40}$ \\ 
				\hline
				\multirow{3}{*}{PEMSD7(L)} 
				& MAE  & $5.31$ & $3.02$ & $3.13$ & $3.10$ & $\underline{2.85}$ & $2.90$ & $3.02$ & $2.97$ & $\underline{2.85}$ & $\mathbf{2.79}$  & $2.89$ & $\mathbf{2.79}$ \\ 
				& RMSE  & $10.07$ & $5.97$ & $6.34$ & $6.07$ & $5.84$ & $5.82$ & $6.26$ & $6.11$ & $5.83$ & $\underline{5.71}$ & $5.85$ & $\mathbf{5.70}$\\
				& MAPE & $13.15$ & $7.64$ & $7.97$ & $7.96$ & $7.24$ & $7.38$ & $7.83$ & $7.40$ & $7.30$ & $\mathbf{7.05}$ & $7.35$ & $\underline{7.09}$ \\ 
				\hline
			\end{tabular}
		}
	\end{table*}
	
	\subsection{Ablation Studies}
\begin{figure}[htb]
		\centerline{\includegraphics[width=\linewidth]{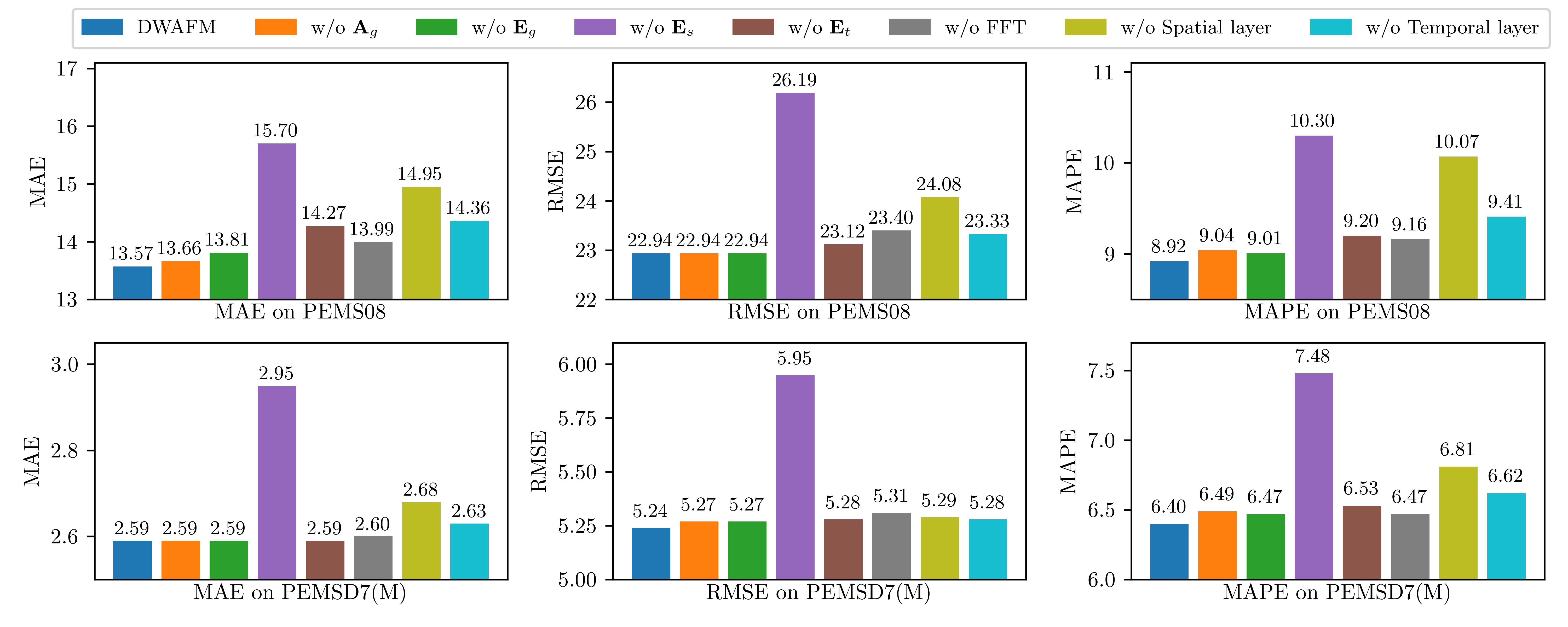}}
		\caption{Ablation study on PEMS08 and PEMSD7(M).}
		\label{fig:3}
	\end{figure}

	To evaluate the performance of each component of our model, we conduct ablation studies on PEMS08 and PEMSD7(M)
	using seven model variants:
	\begin{itemize}
		\item w/o \(\mathbf{A}_g\). It replaces the dynamic weighted adjacency matrix \(\mathbf{A}_g\) with the predefined adjacency matrix \(\mathbf{A}_p\).
		\item w/o \(\mathbf{E}_g\). It removes the DWGS embedding \(\mathbf{E}_g\).
		\item w/o \(\mathbf{E}_s\). It removes the spatial embedding \(\mathbf{E}_s\).
		\item w/o \(\mathbf{E}_t\). It removes the temporal embedding \(\mathbf{E}_t\).
		\item w/o FFT. It removes the FFT and IFFT modules and replaces the temporal layer with a single MLP architecture.
		\item w/o Spatial layer. It removes the spatial layer.
		\item w/o Temporal layer. It removes the temporal layer.
	\end{itemize}
	As shown in Fig.~\ref{fig:3}, all components of the model contribute significantly to improving prediction accuracy. Among embedding approaches, removing either \(\mathbf{E}_g\) or \(\mathbf{E}_s\) results in decreased performance, confirming the effectiveness of learning DWGS via self-attention and integrating them with rich spatial features (e.g., \(\mathbf{E}_a\)) to enhance performance. Notably, removing the FFT module results in a performance drop, highlighting its essential role in capturing temporal patterns.
	
	\subsection{Parameter Sensitivity Analysis}
\begin{figure}[htb]
		\centerline{\includegraphics[width=\linewidth]{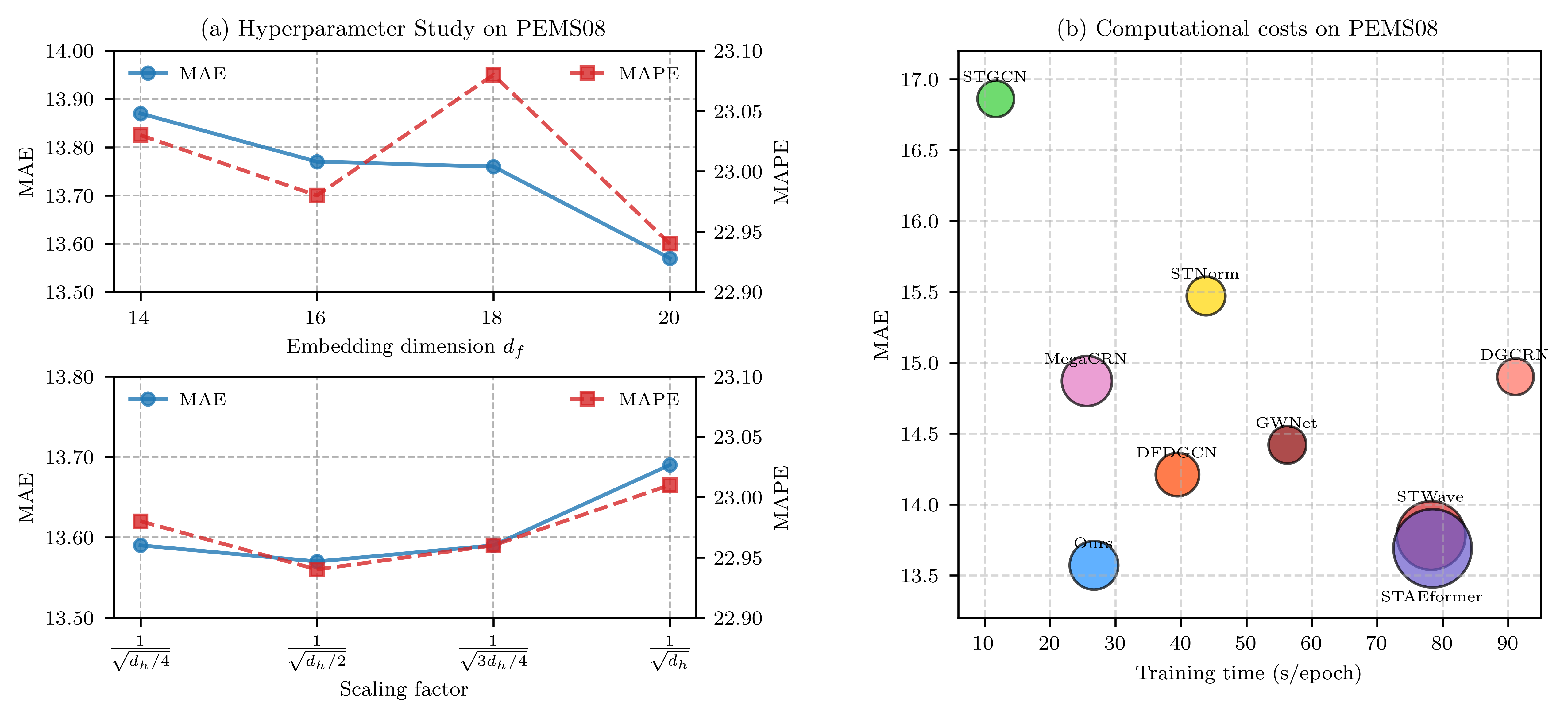}}
		\caption{Hyperparameter study and computational cost analysis on PEMS08.}
		\label{fig:4}
	\end{figure}

	To investigate the impact of core hyperparameters, we evaluate the model sensitivity regarding the embedding dimension $d_f$ and the spatial self-attention scaling factor on the PEMS08 dataset. As illustrated in Fig. \ref{fig:4}(a), the Mean Absolute Error (MAE) generally decreases as $d_f$ increases, reaching its optimum at $d_f = 20$. This suggests that the complex traffic patterns in PEMS08 necessitate a higher-dimensional embedding to capture sufficiently rich latent features. 
	
	Regarding the scaling factor, we deviate from the standard Transformer scaling of $1/\sqrt{d_h}$ to better account for the strong local correlations inherent in traffic data. Experimental results indicate that a scaling factor of $1/\sqrt{d_h/2}$ yields the best performance on PEMS08. This adjustment effectively concentrates the attention mechanism on highly relevant neighboring nodes while suppressing noise from distant, weakly correlated sensors, thereby enhancing the model's capacity for local spatial dependency modeling. Furthermore, parameter sensitivity evaluations on the PEMSD7(M) dataset reveal dataset-specific variations in optimal settings. Additionally, an analysis of the model's sensitivity to node embedding initialization is conducted, with full details and discussions provided in \textit{Appendix B of the Supplementary Material}.
	
	\subsection{Efficiency Study}
	The computational efficiency of the proposed DWAFM is benchmarked against state-of-the-art models on PEMS08, considering training time (seconds per epoch), MAE, and memory consumption (represented by the bubble area). As shown in Fig. \ref{fig:4}(b), DWAFM achieves a superior performance-efficiency trade-off compared to Transformer-based architectures (e.g., STAEformer) and mainstream graph convolutional methods (e.g., DGCRN). Notably, while the computational cost of DWAFM is not the absolute lowest, it achieves substantial performance improvements without introducing excessive overhead, demonstrating an excellent balance between predictive accuracy and practical deployment efficiency. Furthermore, additional efficiency evaluations on PEMS04 are available in \textit{Appendix B of the Supplementary Material}, which further confirm the strong performance-efficiency balance of DWAFM.
	
	\begin{figure}
		\centerline{\includegraphics[width=\linewidth]{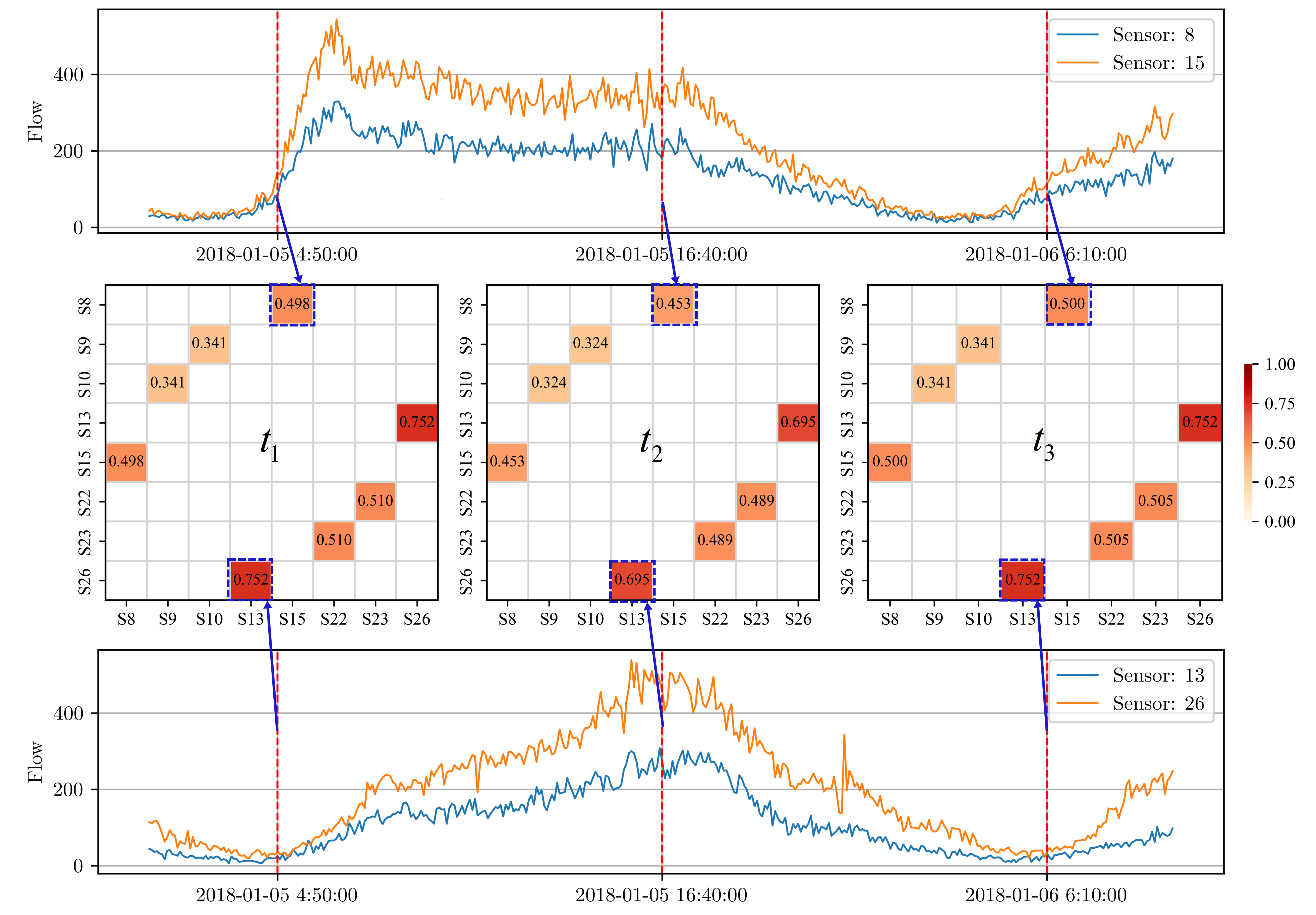}}
		\caption{Visualization of learned dynamic weighted adjacency matrix.}
		\label{fig:5}
	\end{figure}
	
	\subsection{Effectiveness of Temporal Modeling Modules}
	
	To evaluate the proposed frequency-domain MLPs (Fre-MLPs), we compared it with CNN and multi-head self-attention mechanisms. As shown in Table~\ref{tab:2},  Fre-MLPs consistently achieve the lowest errors (MAE, RMSE, and MAPE) across both PEMS08 and PEMSD7(M) datasets. By leveraging frequency-domain transformations, Fre-MLPs effectively capture long-range dependencies and periodic features, overcoming the local receptive field limitations of CNNs and the redundant noise generation often seen in self-attention mechanisms. Furthermore, Fre-MLPs achieve a superior balance between representation accuracy and computational efficiency. While self-attention suffers from high computational complexity (e.g., requiring 23.40s on PEMSD7(M)), Fre-MLPs maintain an execution speed (19.53s) strictly comparable to the lightweight CNN (19.48s), yet deliver significantly enhanced predictive performance.
	
	\begin{table}[h]
		\centering
		\caption{Performance Comparison of Different Temporal Modeling Modules.}
		\label{tab:2}
		\resizebox{\columnwidth}{!}{%
			\begin{tabular}{l|cccc|cccc}
				\hline
				\multirow{2}{*}{Models} & \multicolumn{4}{c|}{Metric (PEMS08)} & \multicolumn{4}{c}{Metric (PEMSD7(M))} \\ \cline{2-9} 
				& MAE & RMSE & MAPE & Time/s & MAE & RMSE & MAPE & Time/s \\ \hline
				CNN & $13.83$ & $23.70$ & $9.16$ & $\mathbf{26.63}$ & $2.60$ & $5.27$ & $6.51$ & $\mathbf{19.48}$ \\
				Attention & $13.85$ & $23.54$ & $9.14$ & $26.85$ & $2.73$ & $5.26$ & $6.91$ & $23.40$ \\ \hline
				\textbf{Fre-MLPs} & $\mathbf{13.57}$ & $\mathbf{22.94}$ & $\mathbf{8.92}$ & $26.71$ & $\mathbf{2.59}$ & $\mathbf{5.24}$ & $\mathbf{6.40}$ & $19.53$ \\ \hline
		\end{tabular}}%
	\end{table}
	
	\subsection{Visualization}
	To validate the effectiveness of the learned \(\mathbf{A}_g\), we visualize and analyze its behavior on PEMS04. We select two sensor pairs (S8 and S15, and S13 and S26) to explore the relationship between the inferred correlation strength and the actual variations in traffic flow. As shown in Fig.~\ref{fig:5}, the correlation strength between S8 and S15 varies consistently with their traffic flow similarity. At time \(t_1\), the highly similar flow values imply strong correlation. By \(t_2\), the flows diverge, leading to noticeable decrease in correlation strength. At \(t_3\), the flows realign, and the correlation recovers accordingly. A similar pattern is observed for sensors S13 and S26. These results demonstrate that the DWGS constructed in this letter effectively captures the dynamic evolution of traffic flow correlations between sensor nodes. The visualization results verify the correlation hypothesis between the strength of the association and the observed data, reflecting that the model has good sensitivity and adaptability to changes in real traffic conditions.
	
	\begin{figure}[hbt]
		\centerline{\includegraphics[width=\linewidth]{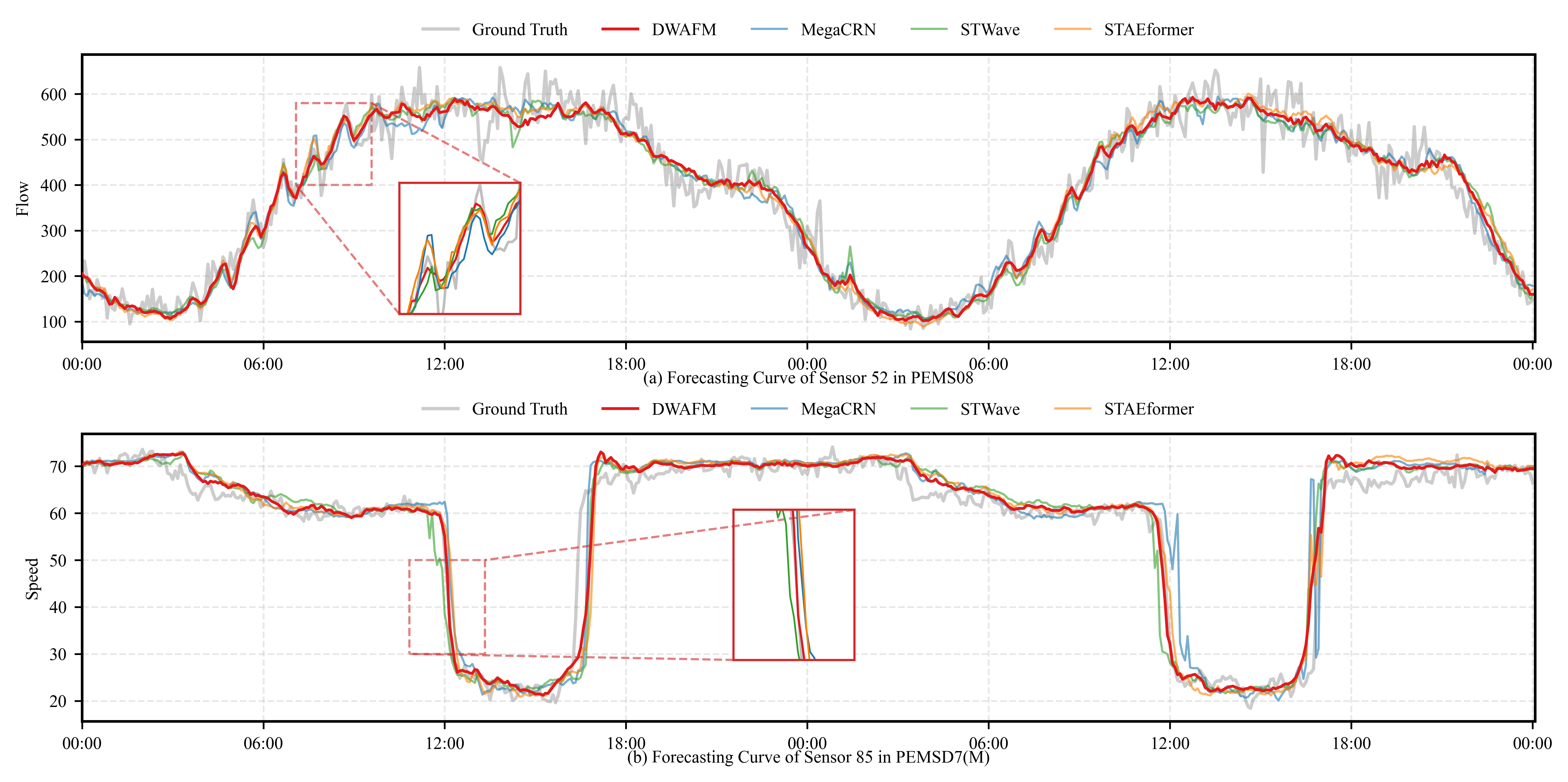}}
		\caption{Visualization of real traffic forecasting results.}
		\label{fig:6}
	\end{figure}
	
	Building upon this effective capacity to capture dynamic spatial dependencies, we further evaluate the final forecasting performance intuitively by visually comparing the model's predictions against the ground truth on the PEMS08 and PEMSD7(M) test sets. Specifically, we select Node 52 of PEMS08 and Node 85 of PEMSD7(M) as representatives to conduct a comparative analysis of the prediction curves between DWAFM and several advanced models, including MegaCRN, STWave, and STAEformer.
	
	As illustrated in Fig.~\ref{fig:6}, the proposed DWAFM exhibits significant superiority in handling complex traffic dynamics. During periods of rising traffic flow and intense fluctuations on PEMS08, DWAFM tracks the changing trends of the ground truth more precisely. Furthermore, in the speed forecasting task on PEMSD7(M), DWAFM consistently outperforms competitive methods such as STAEformer, STWave, and MegaCRN, particularly when facing abrupt speed drops (indicating congestion) as well as during stable high-speed intervals.
	
	\section{Conclusion}
	This study focuses on a core aspect of data representation learning, that is, input embedding techniques. We introduce a novel DWGS embedding approach and present the DWAFM, a predictive model that achieves state-of-the-art performance across five traffic benchmarks. Experimental analysis confirms that the graph structure learned by our embedding method accurately captures real-world, time-varying inter-node relationships, and the DWAFM effectively captures dependencies both within and across sequences. Future work will pursue more concise and efficient architectural designs.
	\newpage
	
	\bibliographystyle{IEEEtran}
	\balance
	\bibliography{reference}
\end{document}

\title{Supplementary Material for ``DWAFM: Dynamic Weighted Graph Structure Embedding Integrated with Attention and Frequency-Domain MLPs for Traffic Forecasting"}
	\author{Sen~Shi, Zhichao~Zhang,~\IEEEmembership{Member,~IEEE}, and Yangfan~He}
	\markboth{}
	{Shell \MakeLowercase{\textit{et al.}}: Bare Demo of IEEEtran.cls for IEEE Journals}
	\maketitle
	\appendices
	\section{Supplementary Details on Methodology and Experiments}
	
	This appendix provides supplementary information regarding our experimental setup and methodology. Specifically, Table~\ref{tab:1} summarizes the statistical characteristics of the evaluated datasets. Algorithm~\ref{alg:1} outlines the complete training procedure and forward propagation data flow of the proposed DWAFM. Furthermore, Table~\ref{tab:2} presents a systematic architectural comparison between DWAFM and existing baselines, clarifying the structural differences in dynamic graph construction and spatio-temporal modeling.
	\begin{table}[h]
		\centering
		\caption{Statistics of Datasets.}
		\renewcommand{\arraystretch}{0.9}
		\setlength{\tabcolsep}{4pt}
		\label{tab:1}
		% \tiny
		\begin{tabular}{ccccc}
			\hline 
			Datasets & Nodes & Length & Sample Rate & Type\\
			\hline
			PEMS08   & $170$  & $17856$ & $5$min & Traffic flow  \\
			
			PEMS04   & $307$  & $16992$ & $5$min & Traffic flow  \\
			
			PEMS03   & $358$  & $26208$ & $5$min & Traffic flow  \\
			\
			PEMSD7(M) & $228$  & $12672$ & $5$min & Traffic speed \\
			
			PEMSD7(L) & $1026$ & $12672$ & $5$min & Traffic speed \\
			\hline
		\end{tabular}
	\end{table}

	\begin{algorithm}[hbt!]
		\caption{The learning algorithm of DWAFM}
		\label{alg:1}
		\begin{algorithmic}[1]
			\State \Input The dataset $\mathbf{X} \in \mathbb{R}^{S \times N \times D}$, pre-defined graph $G=(V, E, \mathbf{A})$, Initialized embedding $f_e(\cdot)$, spatial $f_s(\cdot)$, temporal $f_t(\cdot)$ and regression $f_r(\cdot)$ layers with learnable parameters $\Theta_e, \Theta_s, \Theta_t, \Theta_r$, learning rate $\gamma$, batch size $B$, embedding dimension $d_f$, Number of model layers $m$.
			\State \Set $iter = 1$
			\Repeat
			\State Randomly select a batch (input $\mathbf{X} \in \mathbb{R}^{B \times T \times N \times D}$, label $\mathbf{Y} \in \mathbb{R}^{B \times T_f \times N}$) from $\mathbf{X}$.
			\State Compute $\mathbf{Z} = f_e(\mathbf{X}; \Theta_e) \in \mathbb{R}^{B \times T \times N \times 4d_f}$
			\For{$i = 1$ to $m$}
			\State Compute $\mathbf{Z}_s = f_s(\mathbf{Z}; \Theta_s)$
			\State Compute $\widehat{\mathbf{Z}}_s = \text{LayerNorm}(\mathbf{Z}_s + \mathbf{Z})$
			\State Compute $\mathbf{Z}_t = f_t(\widehat{\mathbf{Z}}_s; \Theta_t) + \mathbf{Z}$
			\EndFor
			\State Compute $\widehat{\mathbf{y}} = f_r(\mathbf{Z}_t; \Theta_r)$
			\State Compute $L = \text{loss}(\widehat{\mathbf{y}}, \mathbf{Y})$
			\State Compute the stochastic gradient of $\Theta_e, \Theta_s, \Theta_t, \Theta_r$ according to $L$.
			\State Update model parameters according to their gradients and the learning rate $\gamma$.
			\State $iter = iter + 1$.
			\Until{convergence}
		\end{algorithmic}
	\end{algorithm}

	\begin{table}[h]
		\caption{Comparison of Graph Construction and Spatial-Temporal Modules Among Different Models.}
		\label{tab:2}
		\centering
		\resizebox{\columnwidth}{!}{%
	 		\begin{tabular}{l|c|c|c|cc}
				\hline
				\toprule
				\multirow{2}{*}{\textbf{Models}} & \textbf{Pre-defined} & \textbf{Graph} & \textbf{S-T} & \multicolumn{2}{c}{\textbf{Core Architecture}} \\ \cline{5-6} 
				& \textbf{Graph} & \textbf{Dynamicity} & \textbf{Embed.} & \textbf{Spatial} & \textbf{Temporal} \\ 
				\hline
				STGCN & $\checkmark$ & Static & - & GCN & CNN \\
				GWNet & $\checkmark$ & Static & - & GCN & CNN \\
				StemGNN & - & Static & - & GCN & CNN \\
				STID & - & - & $\checkmark$ & MLP & MLP \\
				DGCRN & $\checkmark$ & Dynamic & - & GCN & RNN \\
				MegCRN & - & Dynamic & - & GCN & RNN \\
				STwave & $\checkmark$ & Dynamic & - & Attention & CNN + Attention \\
				STAEformer & - & - & $\checkmark$ & Transformer & Transformer \\
				DFDGCN & $\checkmark$ & Static & $\checkmark$ & GCN & CNN \\
				\hline
				\textbf{DWAFM} & $\checkmark$ & \textbf{Dynamic} & $\checkmark$ & \textbf{Attention} & \textbf{Fre-MLPs} \\ 
				\bottomrule
				\hline
			\end{tabular}}%
	\end{table}

	\section{Additional Experimental Results}

	\subsection{Parameter Sensitivity and Efficiency}
\begin{figure}[htb]
		\centerline{\includegraphics[width=\linewidth]{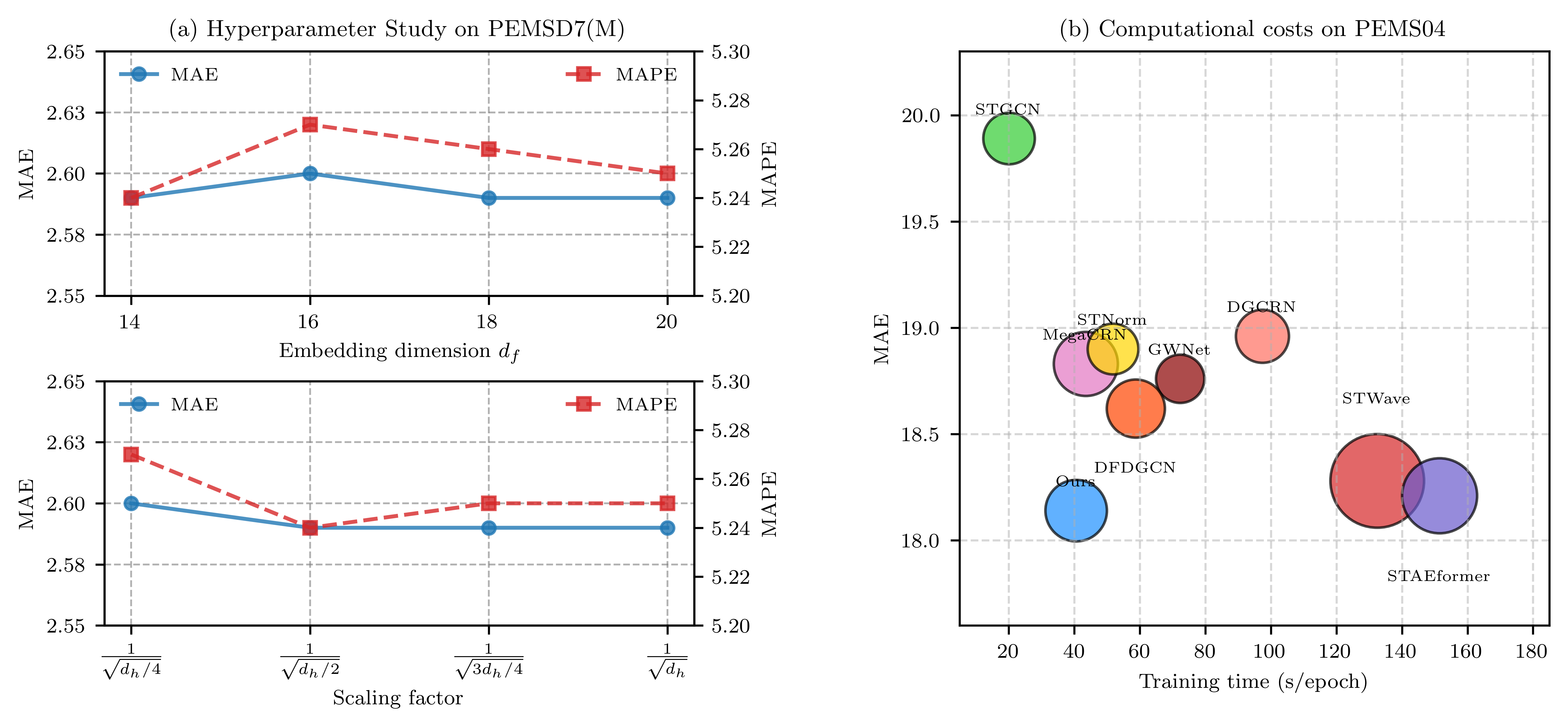}}
		\caption{Hyperparameter study and computational cost analysis on PEMSD7(M) and PEMS04.}
		\label{fig:1}
	\end{figure}

	To investigate hyperparameter sensitivity, we evaluate the embedding dimension $d_f$ and the scaling factor on PEMSD7(M). As shown in Fig.~\ref{fig:1}(a), the model performs best at $d_f=14$, as higher dimensions may cause overfitting on this relatively simple dataset. For the scaling factor, setting it to $1/\sqrt{d_h/2}$ instead of the standard $1/\sqrt{d_h}$ proves optimal. This adjustment effectively emphasizes strong local correlations and suppresses distant noise, thereby enhancing local spatial modeling.
	
	Furthermore, we evaluate the computational efficiency on PEMS04 (Fig.~\ref{fig:1}(b)). Compared to baseline methods like STAEformer and DGCRN, DWAFM achieves substantial accuracy improvements with acceptable training time and memory overhead. This demonstrates that DWAFM maintains an excellent performance-efficiency balance.
	
	\subsection{Sensitivity Analysis of Node Embedding Initialization}
	To rigorously evaluate the model's sensitivity to node embedding initialization, we conducted experiments using four standard techniques: Xavier Uniform, Kaiming Uniform, Orthogonal, and Xavier Normal. As detailed in Table~\ref{tab:3}, the evaluation metrics remain highly stable across all initialization methods on both datasets. Specifically, on PEMSD7(M), the MAE remains constant at 2.59, while the RMSE and MAPE fluctuate only marginally between 5.24--5.25 and 6.40--6.43, respectively. A similar trend is observed on PEMS08, where the MAE and RMSE vary within narrow ranges of 13.54--13.59 and 22.94--23.00, respectively. These consistent results demonstrate that the proposed model exhibits low sensitivity to the initialization schemes of node embeddings, thereby verifying its strong robustness.
	
	\begin{table}[h]
		\centering
		\caption{Performance Comparison under Different Node Embedding Initialization Methods.}
		\label{tab:3}
		\resizebox{\columnwidth}{!}{%
			\begin{tabular}{l|ccc|ccc}
				\hline
				\multirow{2}{*}{Methods} & \multicolumn{3}{c|}{Metric (PEMSD7(M))} & \multicolumn{3}{c}{Metric (PEMS08)} \\ \cline{2-7} 
				& MAE & RMSE & MAPE & MAE & RMSE & MAPE \\ \hline
				Xavier\_uniform & $2.59$ & $5.24$ & $6.40$ & $13.57$ & $22.94$ & $8.92$ \\
				Kaiming\_uniform & $2.59$ & $5.25$ & $6.40$ & $13.59$ & $22.98$ & $8.94$ \\
				Orthogonal & $2.59$ & $5.25$ & $6.43$ & $13.59$ & $23.00$ & $8.92$ \\
				Xavier\_normal & $2.59$ & $5.24$ & $6.42$ & $13.54$ & $22.95$ & $8.90$ \\ \hline
				\textbf{Mean $\pm$ Std} & $\mathbf{2.59 \pm 0.00}$ & $\mathbf{5.25 \pm 0.01}$ & $\mathbf{6.41 \pm 0.02}$ & $\mathbf{13.57 \pm 0.02}$ & $\mathbf{22.97 \pm 0.03}$ & $\mathbf{8.92 \pm 0.02}$ \\ \hline
		\end{tabular}}%
	\end{table}